\documentclass[lettersize,journal]{IEEEtran}

\usepackage{cite}
\usepackage{graphicx}
\usepackage{color}
\usepackage{float}
\usepackage{enumerate}
\usepackage{overpic}
\usepackage[rgb, x11names]{xcolor} 
\usepackage{rotating}
\usepackage{graphicx}
\usepackage{amsthm,amsmath,amssymb}
\usepackage{mathrsfs}
\usepackage{color}
\usepackage{graphics}
\usepackage{multirow}
\usepackage{subfig}
\usepackage{array}
\usepackage{colortbl}
\usepackage{hyperref} 
\usepackage{diagbox}
\usepackage{rotating}
\usepackage{booktabs}

\usepackage{overpic}
\usepackage{contour}
\usepackage{bbding}
\usepackage[marginal]{footmisc}
\usepackage{pifont}
\usepackage{pifont}
\usepackage{booktabs}
\usepackage{makecell}
\usepackage{lineno}
\usepackage{multirow}
\usepackage{marginnote}

\def\ie{\emph{i.e.}}
\def\eg{\emph{e.g.}}
\def\etc{\emph{etc.}}
\def\ourmodel{MCIF-Net}

\newcommand{\myred}[1]{{\textcolor[RGB]{200,30,30}{#1}}}
\newcommand{\myblue}[1]{\textcolor{blue}{#1}}
\newcommand{\mygreen}[1]{\textcolor{Green3}{#1}}

\newcommand{\datasets}{COD} 

\newcommand{\tabincell}[2]{\begin{tabular}{@{}#1@{}}#2\end{tabular}}

\newcommand{\revxrII}[1]{\textcolor{black}{#1}}
\newcommand{\revxrI}[1]{\textcolor{black}{#1}}
\newcommand{\revxr}[1]{\textcolor{black}{#1}}
\newcommand{\revII}[1]{\textcolor{black}{#1}}
\newcommand{\revI}[1]{\textcolor{black}{#1}}

\begin{document}

    
    

\title{Towards Accurate Camouflaged Object Detection with Mixture Convolution and Interactive Fusion}
\author{Geng Chen,~
     Xinrui Chen,~ 
     Bo Dong,~
     Mingchen Zhuge,~
     Yongxiong Wang,~
     Hongbo Bi,~
     Jian Chen,~
     Peng Wang,~
     Yanning Zhang,~\IEEEmembership{Senior Member, IEEE}

\thanks{G. Chen, X. Chen, P. Wang, and Y. Zhang are with National Engineering Laboratory for Integrated Aero-Space-Ground-Ocean Big Data Application Technology, School of Computer Science and Engineering, Northwestern Polytechnical University, China (Emails: geng.chen.cs@gmail.com; xinruichen.cxr@foxmail.com; peng.wang@nwpu.edu.cn; ynzhang@nwpu.edu.cn).}
\thanks{B. Dong is with Center for Brain Imaging Science and Technology, Zhejiang University,  China  (Email: dongbo\_oece@126.com).}
\thanks{M. Zhuge is with School of Computer Science, China University of Geosciences, China (Email: mczhuge@gmail.com).}
\thanks{Y. Wang is with Optical Electrical and Computer Engineering, University of Shanghai for Science and Technology, China (Email: wyxiong@usst.edu.cn).}
\thanks{H. Bi is with School of Electrical Engineering \& Information Department, Northeast Petroleum University, China (Email: bhbdq@126.com).}
\thanks{J. Chen is with  School of Electronic, Electrical Engineering and Physics, Fujian University of Technology, China (Email: jianchen@fjut.edu.cn).}}


\IEEEpubid{}
\maketitle
\IEEEdisplaynontitleabstractindextext
\IEEEpeerreviewmaketitle		
\begin{abstract}
Camouflaged object detection (COD), which aims to identify the objects that conceal themselves into the surroundings, has recently drawn increasing research efforts in the field of computer vision.
In practice, the success of deep learning based COD is mainly determined by two key factors, including (i) A significantly large receptive field, which provides rich context information, and (ii) An effective fusion strategy, which aggregates the rich multi-level features for accurate COD. Motivated by these observations, in this paper, we propose a novel deep learning based COD approach, which integrates the large receptive field and effective feature fusion into a unified framework. Specifically, we first extract multi-level features from a backbone network. The resulting features are then fed to the proposed dual-branch mixture convolution modules, each of which utilizes multiple asymmetric convolutional layers and two dilated convolutional layers to extract rich context features from a large receptive field.
Finally, we fuse the features using specially-designed multi-level interactive fusion modules, each of which employs an attention mechanism along with feature interaction for effective feature fusion.
Our method detects camouflaged objects with an effective fusion strategy, which aggregates the rich context information from a large receptive field. All of these designs meet the requirements of COD well, allowing the accurate detection of camouflaged objects.
Extensive experiments on widely-used benchmark datasets demonstrate that our method is capable of accurately detecting camouflaged objects and outperforms the state-of-the-art methods. 
\end{abstract}


\section{Introduction}\label{sec:intro}
\IEEEPARstart{C}{amouflage} is a biological phenomenon widely existing in nature. Biologists have found that creatures in nature often conceal themselves from predators using their own structures and physiological characteristics.
For instance, a chameleon can change the color of its body when the environment changes;
A crab usually finds the habitat that is similar to its appearance; \etc
These animals camouflage themselves for the survival purpose, such as avoiding being attacked, communication, and courtship.
The earliest research of camouflage can be traced back to the last century.
Thayer \emph{et al.} ~\cite{thayer1918concealing} systematically studied the phenomenon of camouflage in 1918. 
A hundred years have passed, and biologists keep the passion in studying this significant natural phenomenon.
Due to its important scientific and practical value \cite{fan2020Camouflage}, significant efforts have been made to detect/segment the camouflaged objects from natural scenes, which raises the demands of effective camouflaged object detection (COD) approaches.
However, compared with the traditional object detection/segmentation tasks \cite{goferman2011context,long2015fully} in computer vision, COD shows considerable difficulties due to the low boundary contrast between the camouflaged object and its surroundings.
To address this challenge, a number of methods have been proposed to accurately detect/segment camouflaged objects.
For instance, Zhang \emph{et al.}~\cite{zhang2016bayesian} proposed a Bayesian approach to detect moving camouflaged objects.
Deep learning has shown remarkable success in COD and a number of deep-based COD methods have been proposed.
Le \emph{et al.}~\cite{le2019anabranch} designed a general end-to-end network, called anabranch network, for camouflaged object segmentation.
Fan \emph{et al.}~\cite{fan2020Camouflage,fan2021concealed} presented a large COD dataset, called COD10K, and a new COD model, called SINet, which promotes the COD research to a new level.

Despite their advantages, existing deep-based COD methods suffer from two major limitations. First, they usually overlook the importance of large receptive field. In practice, the context information plays an important role in COD, implying that a large receptive is greatly desired since it is able to provide rich context features, which are essential to the accurate detection of camouflaged objects.
Second, most methods fuse the multi-scale features using very simple operations, \eg, concatenation and addition, which are unable to capture the valuable information highly-related to the detection of camouflaged objects. 
This inevitably results in unsatisfactory performance and raises the demands of more advanced feature fusion strategies.

To this end, we propose a novel deep-based COD model, which employs specially-designed mixture convolution and interactive fusion (MCIF) techniques to accurately detect the camouflaged objects from natural scenes.
Our method, called MCIF-Net, harnesses an effective attention-based fusion strategy to aggregate the rich context features extracted from a larger receptive field, which well fits the scenario of accurate COD and overcomes the limitations in existing methods.
In MCIF-Net, we first extract multi-level features from a backbone network.
The resulting features are then fed to our dual-branch mixture convolution (DMC) modules for extracting rich context features from a large receptive field. The DMC module is in a dual-branch form with multiple asymmetric convolutional layers and two dilated convolutional layers, which significantly enlarges the receptive field.
Finally, we propose multi-level interactive fusion (MIF) modules to aggregate the rich context features for the accurate detection of camouflaged objects. Our MIF module employs an attention mechanism along with feature interaction, allowing effective feature fusion.

In a nutshell, the contributions of this paper contain threefold:
\begin{itemize}
	\item We propose a novel feature fusion module, MIF, to effectively aggregates the multi-level features for accurately detecting camouflaged objects. Our MIF module employs an advanced interactive attention mechanism for feature fusion, which guarantees the remarkable performance of MCIF-Net.
	\item We propose an effective receptive field module, DMC, which utilizes multi-type convolution operations to enlarge the receptive field. Our DMC module provides rich context features for COD, boosting the performance significantly.
    \item Extensive experiments are performed using \datasets~benchmark datasets. The experimental results demonstrate that our MCIF-Net outperforms existing cutting-edge models and advances the state-of-the-art performance. In addition, our ablation studies sufficiently demonstrate the effectiveness of two proposed modules, \ie, MIF and DMC.
\end{itemize}

Our paper is organized as follows. In Section~\ref{sec:Related_Works}, we discuss a number of works that are closely related to ours. In Section~\ref{sec:Method}, we provide detailed descriptions for our MCIF-Net and the associated modules. In Section~\ref{sec:Experiments}, we present the implementation details, datasets, evaluation metrics, and experimental results. Finally, we conclude this work and present future directions in Section~\ref{sec:Conclusion}.

\begin{figure*}[t]
	\centering
	{\begin{overpic}[width=0.99\linewidth]
		{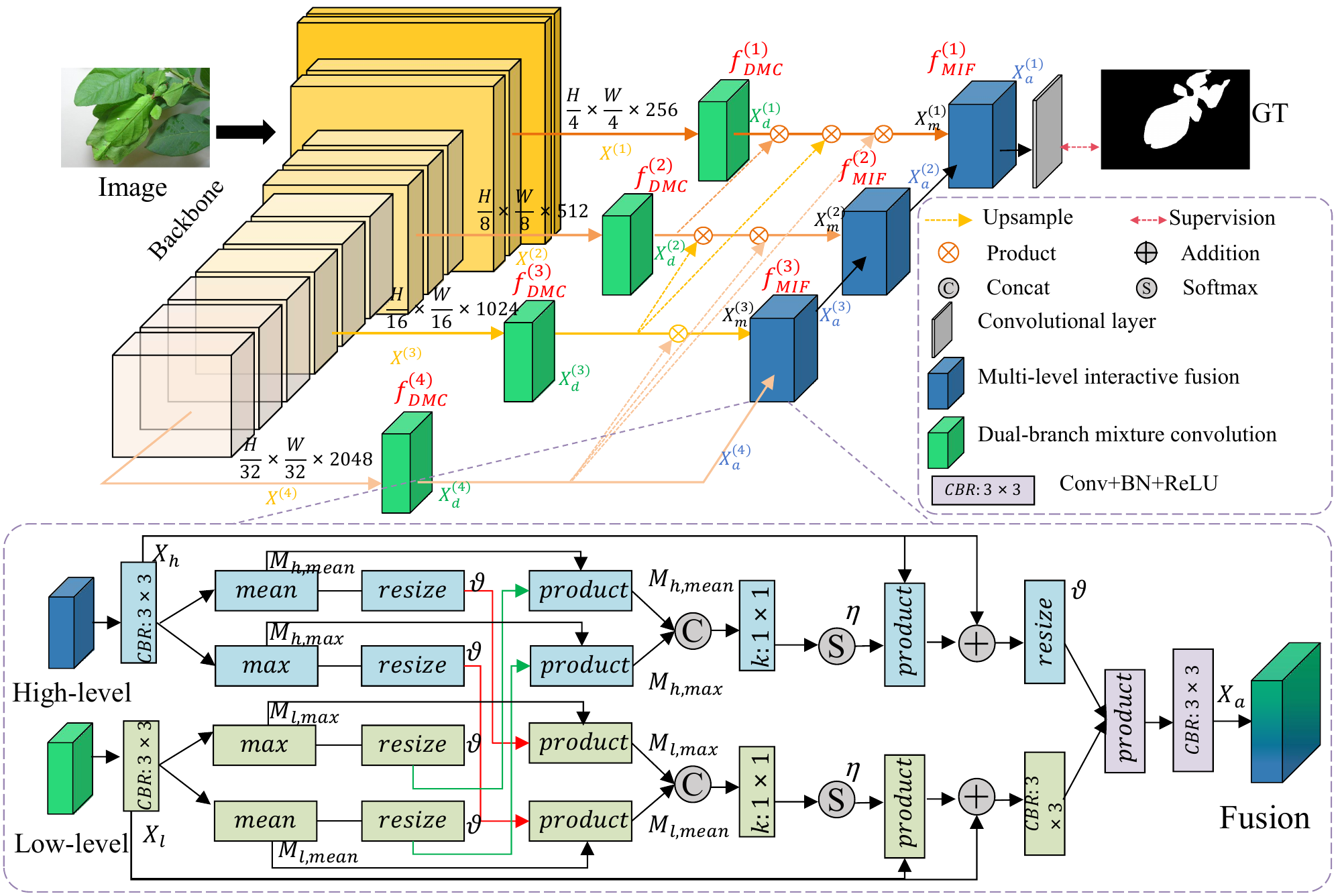}
	\end{overpic}}
	\caption{\revII{An overview of our MCIF-Net. Based on the features from backbone, we first utilize multiple dual-branch mixture convolution modules to extract rich context features and then fuse the features with our multi-level interactive fusion modules for the accurate detection of camouflaged object.}}
	\label{MCIF-Net}
\end{figure*}

\section{Related Works}\label{sec:Related_Works}
\subsection{Camouflage}
Camouflage has been studied for a long time. A large number of biologists explored the typical instances of camouflage and explained camouflage using the principles of natural selection and adaptation.
Inspired by this important natural phenomenon, humans have made attempts to imitate these pattern in many fields, such as art, agriculture, \etc
Camouflage also gained attention from the optical researchers \cite{copeland1998signature}, who proposed the signature strength metrics for camouflaged objects.
Some works have been made to analyze the camouflaged images and to study how to create digital camouflaged images \cite{owens2014camouflaging}.
\revxrII{With the rapid development of object detection \cite{zhai2022deep,yao2021boundary}, COD has drawn increasing research efforts from the computer vision community \cite{lyu2023uedg,mondal2020camouflaged}.}
In particular, Fan \emph{et al.}~\cite{fan2020Camouflage} performed a comprehensive study on COD from the perspective of computer vision and provided an elaborate datasets, \ie, COD10K, which contains 10,000 pictures of 69 categories.
Moreover, efforts have been directed to identifying camouflaged objects using motion information \cite{lamdouar2020betrayed} or bio-inspired adversarial attack \cite{yan2020mirrornet}.
\revII{
MGL~\cite{zhai2021Mutual} decomposes the COD task into object position detection and detail capture.
It repeatedly infers and fuses them through graph convolution.
PFNet~\cite{Mei_2021_CVPR} imitates the predation process, including positioning module and focusing module.
The positioning module is designed to mimic the hunting tactic used to find the potential target object. It uses the global perspective and the focusing module to recognize the object during this hunting process. By focusing on fuzzy information, it gradually narrows down the initial prediction area.
\revxrII{Meanwhile, there are some other novel solutions have been proposed~\cite{fan2021concealed,fan2021ugtr,zhou2024decoupling}.}}

%
\revI{However, accurate COD is a difficult task suffering from three major challenges:
(i) Unlike conventional objects, camouflaged objects share similar appearances with the background. This is due to the fact that animals tend to change their appearances to ``perfectly'' embed into the surrounding environment in order to avoid being identified. Due to the nature of camouflage, it is very challenging the accurately identify camouflaged objects from complex scenes.
(ii) The camouflaged objects are with varied appearances, such as size, shape, \etc, which can reduce the robustness of COD models.
(iii) The wild animals usually live in complex natural environments, implying that the images are usually with complex background, which further aggravates the difficulty of COD.}

\subsection{Attention Mechanism}
Attention mechanism is inspired by the fact that human beings or understanding scenes are not visual analysis of the whole scene, but significant parts.
A number of attention modules have been embedded into the CNN network to improve the understanding ability of the network, which also enables to capture the long-range dependency.
For self-attention mechanisms \cite{wang2018non}, a weighted sum of all positions in spatial temporal domain is calculated as the response at a position.
Squeeze-and-excitation network \cite{hu2018squeeze} formulates channel-wise relationships via an attention-and-gating mechanism.
Spatial and channel-wise attention (SCA) \cite{chen2017sca} utilizes both channel wise attention and spatial attention for image captioning.
Similar to SCA, bottleneck attention module \cite{park2018bam} and convolutional block attention module \cite{woo2018cbam} infer attention maps along two separated dimensions, \ie, channel and spatial, and employs multiple attention maps for adaptive feature refinement.
Fu \emph{et al.}~\cite{fu2019dual} constructed a novel self-attention network consisted of two parallel attention modules, including a position attention module used to obtain the dependency of any two positions in the feature map, and a channel attention module for the dependency between any two channels.
Different from existing attention mechanisms, we propose the MIF module based on an interactive attention mechanism, allowing the effective fusion of multi-level features.

\subsection{Receptive Field Module}
A number of methods have been proposed to study the receptive field in CNN.
Typical instances include the inception block \cite{szegedy2015going}, spatial pyramid pooling (SPP)~\cite{2014Spatial}, atrous spatial pyramid pooling (ASPP)~\cite{chen2017deeplab},and receptive field block (RFB)~\cite{liu2018receptive}.
Inception block \cite{szegedy2015going} consists of multiple branches, where the standard convolutional layers with different kernel sizes are employed to extract multi-level features. The resulting features are then combined as the output, which encodes the rich representations from a large receptive filed. 
However, inception block may loss some crucial details since all the kernels are used at the same center.
Different from the inception block, SPP \cite{2014Spatial} generates a fixed-length feature vector from the feature maps with different scales, and transmitted it to the fully connected layer through pooling of three scales.
ASPP \cite{chen2017deeplab} introduces dilated convolution for enlarged receptive field, however it uses same kernel size, which may lead to a confusion between context and object.
\revII{GCN \cite{peng2017large} enlarges the receptive field by considering large convolutional kernels.}
To resolve these limitations, RFB \cite{liu2018receptive} assigns larger weights to the positions nearer to the center, providing improved performance.

\section{Method}\label{sec:Method}	
An overview of our MCIF-Net is shown in Fig.~\ref{MCIF-Net}. In general, MCIF-Net consists of a Res2Net-50 based backbone and two kinds of key modules, \ie, dual-branch mixture convolution (DMC) module and multi-level interactive fusion (MIF) module. In what follows, we will first present the overall architecture of MCIF-Net in Section~\ref{Network}. We will then detail in two proposed key modules, \ie, DMC (Section~\ref{DMC}) and MIF (Section~\ref{CSM}). Finally, we will describe our loss function in Section~\ref{loss}.

\subsection{Network Architecture}\label{Network}	
As shown in Fig.~\ref{MCIF-Net}, given a RGB image $\mathbf{X}\in{\mathbb{R}^{W\times H\times 3}}$, where $H$ and $W$ denote the image width and height, respectively, we first extract four levels of hierarchical features \revI{$\mathbf{X}^{(i)} \in \mathbb{R}^{\frac{H}{2^{i+1}} \times \frac{W}{2^{i+1}} \times C_i}$} using a backbone network, \revI{where $i \in \{1,2,3,4\}$ and $C_i \in \{256,512,1024,2048\}$}. \revI{However, these features are unable to capture the rich context information due to the use of convolution, which is a local operation providing a relatively small receptive field.}
To resolve this issue, we feed each $\mathbf{X}^{(i)}$ to the corresponding DMC module \revI{to enlarge the receptive field for extracting richer context features.}
The resulting features \revI{$\mathbf{X}_\text{d}^{(i)} \in \mathbb{R}^{\frac{H}{2^{i+1}} \times \frac{W}{2^{i+1}} \times 64}$} are defined as
\begin{equation}
\mathbf{X}_\text{d}^{(i)} = \mathit f_{\text{DMC}}^{(i)}({\mathbf{X}^{(i)}}),
\end{equation} 
where $f_{\text{DMC}}^{(i)}(\cdot)$ denotes a function acting as the $i$th DMC module.

\revII{To explore the correlations between the features from adjacent layers, we multiply them element-wisely, resulting new features $\mathbf{X}^{(j)}_{\text{m}}$. Mathematically, we define $\mathbf{X}^{(j)}_{\text{m}}$ as
\begin{equation}
\mathbf{X}^{(j)}_{\text{m}} = \mathbf{X}_\text{d}^{(j)} \odot \prod_{k=j+1}^{L}Conv(up(\mathbf{X}_\text{d}^{(k)})),
\end{equation} 
Where $Conv(\cdot)$ is a convolution layer, $up(\cdot)$ denotes the upsample operation, $\odot$ denotes the Hadamard product, and $j \in \{1,2,3\}$. Since there are four side output levels, we set $L = 4$.}
\revI{After that, we fuse the multi-level features using our MIF modules, which are designed based on an interactive attention mechanism.} As shown in Fig.~\ref{MCIF-Net}, three MIF modules are adopted, where each MIF has two inputs, including multiplication features $\mathbf{X}_{\text{m}}^{(j)}$ and the features from the last MIF, \ie,
\begin{equation}
\mathbf{X}_\text{a}^{(j)} = f_{\text{MIF}}^{(j)}(\mathbf{X}_{\text{m}}^{(j)},\mathbf{X}_\text{a}^{(j+1)}),
\end{equation}
where \revI{$\mathbf{X}_\text{a}^{(j)} \in \mathbb{R}^{\frac{H}{2^{j+1}} \times \frac{W}{2^{j+1}} \times 64}$} is the output of $j$th MIF module, and $f_{\text{MIF}}^{(j)}(\cdot)$ denotes a function acting as the $j$th MIF module. Note that $\mathbf{X}_\text{a}^{(4)}$ denotes the features provided by the fourth DMC module, \ie, $\mathbf{X}_\text{a}^{(4)} = \mathbf{X}_\text{d}^{(4)}$.

Finally, we feed \revI{$\mathbf{X}_\text{a}^{(1)} \in \mathbb{R}^{\frac{H}{4} \times \frac{W}{4} \times 64}$} to a convolutional layer \revI{to adjust the number of channels to one and then upsample it to match the size of input for getting the final prediction $P^{(1)}\in \mathbb{R}^{H \times W \times 1}$}, which is a map encoding the pixel-wise probabilities of camouflaged objects. 
\begin{figure}[t]
	\centering
	\begin{overpic}[width=1\linewidth]{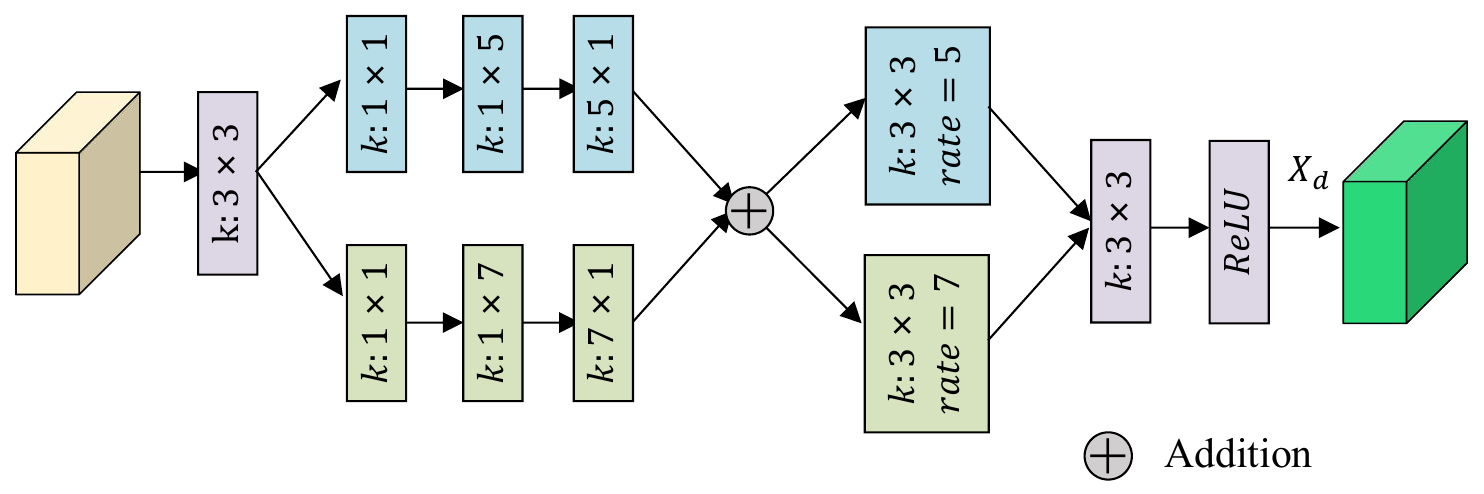}
	\end{overpic}
	\vspace{-18pt}
	\caption{Illustration of our dual-branch mixture convolution module. We enlarge the receptive filed using asymmetric convolutional and dilated convolutional layers.}
	\label{MC}
\end{figure}

\subsection{Dual-branch Mixture Convolution Module}\label{DMC}
To enlarge the receptive filed, we propose a novel receptive filed module, DMC, which is a stack of asymmetric convolutional and dilated convolutional layers. As shown in~\ref{MC}, we employ the DMC to extract rich context features on each side output of the backbone network. Specifically, to learn equally-spatial-sized features with a significantly larger receptive field, the resulting features are then projected to two independent branches. Each of them is equipped with a convolutional layer with $kernel=1\times1$ for changing the number of channels, followed by a pair of asymmetric convolutional layers with $kernel=1\times n$ and $kernel=n \times 1$ for enlarging the receptive field, where $n=5$ for the top branch and $n=7$ for the bottom branch.
{To maximize the co-characterization between two branches}, we add the features from two branches for the superimposed features, \ie,
\begin{equation}
\mathbf{\hat{X}}^{(i)} = \mathbf{X}^{(i)}_{\text{TB}} + \mathbf{X}^{(i)}_{\text{BB}},\\
\end{equation}
where \revI{$\mathbf{X}^{(i)}_{\text{TB}}\in \mathbb{R}^{\frac{H}{2^{i+1}} \times \frac{W}{2^{i+1}} \times 64}$} and \revI{$\mathbf{X}^{(i)}_{\text{BB}}\in \mathbb{R}^{\frac{H}{2^{i+1}} \times \frac{W}{2^{i+1}} \times 64}$} denote the features from top branch and bottom branch, respectively. 

To further enlarge the receptive field, we project features to two branches equipped with the dilated convolutional layers \cite{yu2015multi} with the same kernel size ($kernel=3\times3$) and different rates ($rate=5$ for the top branch and $rate=7$ for the bottom branch).
Finally, we merge again using the element-wise addition and utilize a convolutional layer with $kernel=3\times3$ followed by a ReLU activation function \cite{glorot2011deep} for the rich context features $\mathbf{X}_\text{d}^{(i)}$, \ie,
\begin{equation}
\mathbf{X}_\text{d}^{(i)} = CR({DC_{r = 5}}(\mathbf{\hat{X}}^{(i)}) + {DC_{r = 7}}(\mathbf{\hat{X}}^{(i)})),
\end{equation}
where $DC_{r = 5}(\cdot)$ and $DC_{r = 7}(\cdot)$ represent the dilated convolutional layers with different rates, $r$, and $CR(\cdot)$ denotes the combination of the final convolutional layer and a ReLU activation function.


\subsection{Multi-level Interactive Fusion Module}\label{CSM}	
\revI{In practice, high-level features provide rich semantic information, while low-level features contain more detailed information.
Therefore, we propose an effective multi-level feature fusion module, MIF, which employs an interactive attention mechanism to gradually integrate high-level and low-level features in a top-down manner.}
Unlike existing feature fusion strategies, which usually rely on simple feature addition (\eg, EGNet \cite{zhao2019egnet}) and concatenation (\eg, UNet++ \cite{zou2018DLMIA}), our MIF module fuses the multi-level features using an attention mechanism along with feature interaction, allowing effective feature fusion for the accurate detection of camouflaged objects.

Fig.~\ref{MCIF-Net} illustrates the architecture of MIF module.
In order to better illustrate the feature scale change, we take middle MIF module as an example.
Denote the low and high-level features as
\revI{$\mathbf{X}_{\text{l}} \in{\mathbb{R}^{\frac{H}{8}\times \frac{W}{8}\times 64}}$} and \revI{$\mathbf{X}_{\text{h}}\in{\mathbb{R}^{\frac{H}{16}\times \frac{W}{16}\times 64}}$}.
We first utilize two symmetric branches to process two levels of features and then employ the features in two branches to enhance each other.
Using the low-level branch as an example, we feed
$\mathbf{X}_{\text{l}}$ 
to a convolution block and have new features denoted as
$CBR(\mathbf{X}_{\text{l}})$, where 
$CBR(\cdot)$ 
denotes a convolution block with a convolutional layer
($kernel=3\times3$) 
followed by a batch normalization layer and a ReLU activation function.
At each spatial location, we then computed the statistics, \ie, the average and maximum, across the channel dimension.
Mathematically, the resulting informative spatial attention maps 
\revI{$M_{\text{l,max}}\in{\mathbb{R}^{\frac{H}{8}\times \frac{W}{8}\times 1}}$} and \revI{$M_{\text{l,mean}}\in{\mathbb{R}^{\frac{H}{8}\times \frac{W}{8}\times 1}}$}
are defined as
\begin{align}M_{\text{l,max}} &= \sigma(max_{\text{c}}(CBR(\mathbf{X}_{\text{l}}))),\\
M_{\text{l,mean}} &= \sigma(mean_{\text{c}}(CBR(\mathbf{X}_{\text{l}}))),
\end{align}
where $max_{\text{c}}(\cdot)$ and $mean_{\text{c}}(\cdot) $ denote two respective operations for computing max and mean maps across the channel dimension, and $\sigma(\cdot)$ denotes the Sigmoid activation function.
Similarly, for the high-level branch, we have \revI{$M_{\text{h,max}}\in{\mathbb{R}^{\frac{H}{16}\times \frac{W}{16}\times 1}}$} and \revI{$M_{\text{h,mean}}\in{\mathbb{R}^{\frac{H}{16}\times \frac{W}{16}\times 1}}$}, which are generated using $\mathbf{X}_{\text{h}}$ with the same procedure.

After that, we enhance the attention maps form two branches. For the low-level branch, the enhancement process is described as follows
\begin{equation}
M_{\text{l,max}} = mul(M_{\text{l,max}},\vartheta (M_{\text{h,mean}})),
\end{equation}
\begin{equation}
M_{\text{l,mean}} = mul(M_{\text{l,mean}},\vartheta (M_{\text{h,max}}),
\end{equation}
where \revI{$M_{\text{l,max}}\in{\mathbb{R}^{\frac{H}{8}\times \frac{W}{8}\times 1}}$} and \revI{$M_{\text{l,mean}}\in{\mathbb{R}^{\frac{H}{8}\times \frac{W}{8}\times 1}}$}. The $mul(\cdot)$ denotes the product operation, $\vartheta(\cdot)$ denotes the resizing operation used to match the dimensions of low and high-level attention maps. Similarly, for the high-level branch, we compute the enhanced attention maps \revI{$M_{\text{h,max}}\in{\mathbb{R}^{\frac{H}{16}\times \frac{W}{16}\times 1}}$} and \revI{$M_{\text{h,mean}}\in{\mathbb{R}^{\frac{H}{16}\times \frac{W}{16}\times 1}}$} as
\begin{equation}
M_{\text{h,max}} = mul(M_{\text{h,max}},\vartheta (M_{\text{l,mean}})),
\end{equation}
\begin{equation}
M_{\text{h,mean}} = mul(M_{\text{h,mean}},\vartheta (M_{\text{l,max}})).
\end{equation}

We then concatenate the attention maps 
$M_{\text{l,mean}}$ 
and
$M_{\text{l,max}}$ 
together and then feed the concatenated map to a convolutional layer with 
$kernel=1\times1$ to reduce the channels to 1, leading to \revI{$M_{\text{l}}\in{\mathbb{R}^{\frac{H}{8}\times \frac{W}{8}\times 1}}$}. 
We do the same for
$M_{\text{h,max}}$ 
and
$M_{\text{h,mean}}$ 
and have
\revI{$M_{\text{h}}\in{\mathbb{R}^{\frac{H}{16}\times \frac{W}{16}\times 1}}$}.
After that, we pass 
$M_{\text{l}}$ 
and 
$M_{\text{h}}$
to two Sigmoid layers
$\eta(\cdot)$ 
and enhance the original features by multiplying and adding element by element, which are defined as follows:
\begin{align}
\mathbf{\hat{X}}_{\text{l}} &= mul(\eta(M_{\text{l}}), {\mathbf{X}_{\text{l}}}) + {\mathbf{X}_{\text{l}}},\\
\mathbf{\hat{X}}_{\text{h}} &= mul(\eta(M_{\text{h}}), {\mathbf{X}_{\text{h}}}) + {\mathbf{X}_{\text{h}}}.
\end{align}

Finally, we upsample \revI{$\mathbf{\hat{X}}_{\text{h}}\in{\mathbb{R}^{\frac{H}{16}\times \frac{W}{16}\times 64}}$} to the same size of $\mathbf{\hat{X}}_{\text{l}}$,and feed \revI{$\mathbf{\hat{X}}_{\text{l}}\in{\mathbb{R}^{\frac{H}{8}\times \frac{W}{8}\times 64}}$} to a convolutional block so that the resulting low and high-level features, $\vartheta({\mathbf{\hat{X}}_{\text{h}}})$ and $CBR({\mathbf{\hat{X}}_{\text{l}}})$, share the same dimensions. Finally, we perform element-wise multiplication and feed the features to the last convolutional block for the final enhanced features \revI{$\mathbf{X}_{\text{a}}\in{\mathbb{R}^{\frac{H}{8}\times \frac{W}{8}\times 64}}$}. Mathematically, $\mathbf{X}_{\text{a}}$ is computed using
\begin{equation}
\mathbf{X}_{\text{a}} = {CBR}(mul(\vartheta({\mathbf{\hat{X}}_{\text{h}}}), CBR({\mathbf{\hat{X}}_{\text{l}}}))).
\end{equation}

\revII{Different levels of features provide different kinds of information. The high-level features are sensitive to the spatial position information of the object, while the low-level features contain more spatial details. The fusion of the two kinds of features can promote the model to learn more robust features. For the pixels in each spatial position, we first calculate the mean value of the maximum value in the channel dimension. The maximum value helps to find the location information, while the mean value helps to preserve the edge details. Then the high-level and low-level features are replaced in an interactive way. Compared with the original features, the replaced features have been improved in position and detail. Finally, the two are concatenated to further obtain more stable features.}

\subsection{Loss Function}\label{loss}	
The binary cross-entropy (BCE) loss is widely used in binary segmentation.
However, it is only defined in a shallow pixel level, which leads to unsatisfactory performance when there is imbalanced problem.
The intersection-over-union (IoU) loss proposed in BASNet \cite{qin2019basnet} resolves the limitation in BCE by putting more attention on the regional level.
Therefore, we combine these two loss functions to design ours.
In addition, we notice the category imbalance issue of positive (foreground pixel) and negative (background pixel) samples in the training dataset.
To resolve this issue, we add an additional balance parameter for each pixel, termed as $\lambda_{n}$, which is defined as
\begin{equation}
\lambda_n  = \sigma(|P_n - G_n|),
\end{equation}
where $G_n$ and $P_n$ are the values at pixel $n$ in our prediction $P$ and the ground-truth label $G$, respectively.
We then define the improved BCE loss function $\mathcal{L}_{\text{BCE}}(P,G)$ as
\begin{equation}
\mathcal{L}_{\text{BCE}}(P,G)=
-\sum\nolimits_{n = 1}^N [\lambda_n*(G_n \log (P_n)+(1-G_n) \log (1-P_n))],
\end{equation}
where $N=H \times W$ is the total number of pixels.
In addition, we use IoU loss function $\mathcal{L}_{\text{IoU}}(P,G)$ to focus on regions.
The IoU loss is defined as
\begin{equation}\mathcal{L}_{\text{IoU}}(P,G) = 1 - \frac{{\sum\nolimits_{n = 1}^N {[{P_n} \times {G_n}]} }}{{\sum\nolimits_{n = 1}^N {[{P_n}+{G_n} - {P_n} \times {G_n}]} }},
\end{equation}

\revII{Finally, we provide supervisions for the final MIF module and define the total loss function $\mathcal{L}$ as
\begin{equation}
\mathcal{L}=\mathcal{L}_{\text{BCE}}(P,G)+\mathcal{L}_{\text{IoU}}(P,G),
\end{equation}
}

\section{Experiments}\label{sec:Experiments}

\subsection{Implementation Details}\label{sec:Details}
The proposed MCIF-Net is implemented using PyTorch and is publicly available at \revxr{\url{https://github.com/xingshui2021/MCIFNet}}.
\revII{We adopt Res2Net-50 as our backbone and only keep the feature extraction part, which is initialized by the pre-trained model on ImageNet.}
We follow \cite{fan2020Camouflage} to resize the input images to $352 \times 352$ during both training and testing.
We follow the experimental setting in \cite{fan2020Camouflage} and utilize the same dataset for training and testing.
\revxr{The network is trained using an SGD optimizer with an initial learning rate of 0.005.
The batch size is set to 36.
The training takes nearly 4.5 hours for 100 epochs.
The test speed reaches 28.02 Fps on a server equipped with a RTX 3090 GPU.}

\subsection{Datasets}\label{sec:Datasets}
\revII{To evaluate our method thoroughly, we perform extensive experiments on on \datasets~benchmark datasets, including  CAMO \cite{le2019anabranch} and COD10K \cite{fan2020Camouflage}, and NC4K~\cite{yunqiu_cod21}.
}
CAMO is the first formal COD dataset, where each image contains more than one camouflaged objects with pixel-level labels.
The COD10K is the largest COD dataset with pixel-level annotations, which is composed of four main categories {(Amphibian, Acoustic, Flying, and Territorial)} and 69 sub-classes. The Amphibian contains 124 pictures with complex background, which are mainly focused on small targets. The Aquatic contains 474 pictures of camouflaged objects in the water mainly with slender trunks and complex edges. The Flying contains 714 pictures, which are mainly focused on small targets and partially occluded objects.
The Terrestrial is composed of 699 pictures, which contains abundant land creatures with small structure and slender limbs. 
\revII{NC4K~\cite{yunqiu_cod21} is the latest COD dataset, which contains 4,121 images from the Internet. All the image are with fine-grained per-pixel labels. The NC4K is widely used in the testing phase for evaluating the generalization ability of COD models.}

\subsection{Evaluation Criteria}\label{Evaluation}
The COD belongs to the task of binary segmentation, therefore we perform quantitative evaluations using a number of popular binary segmentation evaluation criteria, including S-measure \cite{fan2017structure}, E-measure \cite{fan2018enhanced}, weighted F-measure, and mean absolute error.

\textbf{{S-measure ($S_{\alpha}$)}}. We use S-measure to calculate the spatial structure similarity between {the prediction and the ground-truth in the object-aware level $S_{o}$ and the region-aware level $S_{r}$.} Mathematically, the S-measure $S_{\alpha}$ is defined as
 \begin{equation}
S_{\alpha}=\alpha S_{o}+(1-\alpha) S_{r},
 \end{equation}
where $\alpha$ is a tuning parameter and is set to $0.5$ according to \cite{fan2017structure}.

\textbf{E-measure ($E_\phi$)}. As a binary map evaluation metric, E-measure is designed for evaluating the difference between the predicted map $P$ and ground-truth label $G$ from the local and global perspectives. It is defined as
 \begin{equation}
E_\phi=\frac{1}{W \times H} \sum_{x=1}^{W} \sum_{y=1}^{H} \phi(P(x,y) - G(x,y)),
 \label{eq:Em}
 \end{equation}
where $\phi(\cdot)$ denotes the enhanced alignment matrix.
\revxr{In the paper, we present the maximum E-measure, and likewise, the other COD methods are configured in a similar manner.}

\textbf{Weighted F-measure ($F^{w}_{\beta}$)}. As an improved version of F-measure, $F^{w}_{\beta}$ is defined based on $\text{Recall}$ and $\text{Precision}$ to avoid the influence of threshold, \ie,
 \begin{equation}
F^{w}_{\beta}=\frac{\left(1+\beta^{2}\right)  \text{Precision}^{w} \times  \text{Recall}^{w}}{\beta^{2} \times  \text{Precision}^{w}+ \text{Recall}^{w}},
 \end{equation}
where $\beta^{2}$ is a tuning parameter and is set to 0.3 according to \cite{margolin2014evaluate,fan2020Camouflage}.

\textbf{Mean absolute error (MAE, $M$)}. We employ MAE to measure the average absolute distance between the normalized predicted map and the ground-truth. Specifically, MAE is defined as
 \begin{equation}
M=\frac{1}{W \times H} \sum_{x=1}^{W} \sum_{y=1}^{H}\left|P(x, y) - G(x, y)\right|.
 \end{equation}

\begin{table*}[t!]
	\caption{\revII{\small Quantitative results on COD benchmark datasets. \revxr{The best, second best, and third best results are marked by \myred{red}, \myblue{blue}, and \mygreen{green} colors, respectively.} All methods are trained using the same dataset, as in \cite{fan2020Camouflage}. $\uparrow$ indicates the higher the score the better, and vice versa for $\downarrow$. We evaluate the results using four widely used metrics, including S-measure ($S_\alpha$), E-measure ($E_\phi$), F-measure ($F_\beta$), and MAE ($M$).}}
	
	\centering
	\renewcommand{\arraystretch}{1.35}
	\setlength\tabcolsep{3pt}
	\resizebox{1\textwidth}{!}{
        \begin{tabular}{lc|ccccccccccccc>{\columncolor{gray!30}}c}
			
			\toprule[1.5pt]
			&    Metric   & {{SINet}}     & {{ PraNet}}      & {{ C2FNet}}      & {{UGTR}}      & {{PFNet}}   &{{S-MGL}}  &{{R-MGL}} &{{LSR}} &{{JCSOD}} &{{ERRNet}} &{{SINetV2}} &{{BSANet}} &{{FLCNet}}  & {{MCIF-Net}}  \\
			\midrule
			\multirow{4}[0]{*}{\begin{sideways}\textit{COD10K~\cite{fan2020Camouflage}}\end{sideways}} & $S_{\alpha}\uparrow$     & 0.771    & 0.789      & 0.813     & \myblue{0.818}  & 0.800 &0.811 &0.814 &0.804 &0,809 &0.786 &\mygreen{0.815} & \myblue{0.818}    & \myblue{0.818}      &\myred{0.822} \\
			& $E_\phi\uparrow$     & 0.868     & 0.879     & \mygreen{0.900}    & 0.891 & 0.890 & 0.890 & 0.890 &0.892  & 0.891 &0.886 &\myblue{0.906} &- &\mygreen{0.900}         &\myred{0.908} \\
			& $F_{\beta}\uparrow$      & 0.551    & 0.629     & {{0.686}}    & 0.667 & 0.660 &0.655 &0.666 &0.673 &0.684 &0.630 &0.680 &\myblue{0.699} &\myred{0.700}          &{\mygreen{0.698}} \\
			& $M\downarrow$     & 0.051     & 0.045     &\mygreen{0.036}   & \myblue{0.035} &{{ 0.040}} &0.037 &\myblue{0.035}    &0.037 &\myblue{0.035} &0.043 &0.037 &\myred{0.034}  &\myred{0.034} &\myred{0.034}    \\
			\midrule
			\midrule
			\multirow{4}[0]{*}{\begin{sideways}\textit{CAMO~\cite{le2019anabranch}}\end{sideways}} & $S_{\alpha}\uparrow$      & 0.751     & 0.769    & 0.796     & 0.785 & {{0.782}}  &0.772 &0.775 &0.787 &0.800 &0.779 & \myblue{0.820} &0.796 &\mygreen{0.808} &\myred{0.823} \\
			& $E_\phi\uparrow$     & 0.831     & 0.837    & 0.864     & 0.854 &{{0.855}} &0.842 &0.842 &0.854 &0.873 &0.858 &\myred{0.895} &- & \mygreen{0.882}         &\myblue{0.893} \\
			& $F_{\beta}\uparrow$      & 0.606     & 0.663     & 0.719  & 0.686   & {{0.695}}     &0.664  &0.673 &0.696 &0.728 &0.679 &\myblue{0.743}  &0.717  &\mygreen{0.741}  &\myred{0.748}\\
			& $M\downarrow$     & 0.100      & 0.094     & {{0.080}}    & 0.086 & 0.085 &0.089 &0.088 &0.080 &\mygreen{0.073}  &0.085 &\myred{0.070} &0.079  &\myblue{0.071}        &\myblue{0.071}\\
             \midrule
			\midrule
			\multirow{4}[0]{*}{\begin{sideways}\textit{NC4K~\cite{yunqiu_cod21}}\end{sideways}} & $S_{\alpha}\uparrow$      & 0.808     & 0.822     & 0.838     & 0.839 & {{0.829 }} &0.829 &0.833 &0.840 &\mygreen{0.842} &0.827 &\myred{0.847}   &-  &\myblue{0.845}       &{\myred{0.847}} \\
   
			& $E_\phi\uparrow$     & 0.883     & 0.888    & 0.904     & 0.899  &0.898 &0.893 & 0.893 &0.907 &0.907 &0.901 &\myred{0.914}  &-  &\mygreen{0.911}  &{\myblue{0.912}} \\
			& $F_{\beta}\uparrow$      & 0.723    & 0.724     & 0.762  & 0.747    & 0.745 &0.731 &0.740 &\mygreen{0.766}  &{\myblue{0.770}}    &0.737  &{\myblue{0.770}}  &-   &\myred{0.780}  &{\myblue{0.770}} \\
			& $M\downarrow$     & 0.058     & 0.059   & {{0.049}}    & 0.052 & {{0.053}}  & 0.055 & 0.052 &\mygreen{0.048} &\myblue{0.047} & 0.054&\mygreen{0.048}  &- &\myred{0.046}       &{\myblue{0.047}} \\

			\toprule[1.5pt]
	\end{tabular}}
        \label{all_data}
\end{table*}

\subsection{Comparison with SOTAs}\label{sec:Comparison}
We compare our MCIF-Net with 13 state-of-the-art (SOTA) COD models to validate its effectiveness. These models include SINet~\cite{fan2020Camouflage}, PraNet~\cite{fan2020pra}, C2FNet~\cite{chen2022camouflaged}, UGTR~\cite{fan2021ugtr}, PFNet~\cite{Mei_2021_CVPR},     S-MGL~\cite{zhai2021Mutual}, R-MGL~\cite{zhai2021Mutual}, LSR~\cite{yunqiu_cod21}, JCSOD~\cite{aixuan_cod_sod21}, 
ERRNet~\cite{ji2022fast},
SINetV2~\cite{fan2021concealed}, 
BSANet~\cite{zhu2022can},
FLCNet~\cite{electronics12122570}
 .
We follow the benchmark presented in \cite{fan2021concealed}, where the aforementioned baseline models are trained using the their open source codes with the same dataset as in ours and the default parameter settings suggested in the literature.

\textbf{Quantitative evaluation}.
The quantitative results, shown in in Table~\ref{all_data}, indicate that our MCIF-Net outperforms outperforms SOTA models in terms of a majority of evaluation metrics and consistently achieves promising scores across distinct datasets: NC4K, CAMO,  and COD10K.
This sufficiently demonstrates its robustness and reliability in various testing scenarios.
Compared with the the second best model (\ie, FLCNet), our model significantly improves the S-measure $S_{\alpha}$ in all datasets.
Compared with the edge-guided models (\eg, S-MGL, R-MGL), our model shows improved performance in the absence of additional auxiliary edge guidance, which further demonstrates the effectiveness of our model and its potentials.
%
\begin{figure*}[!ht]
	\centering
	\begin{overpic}[width=1\linewidth]{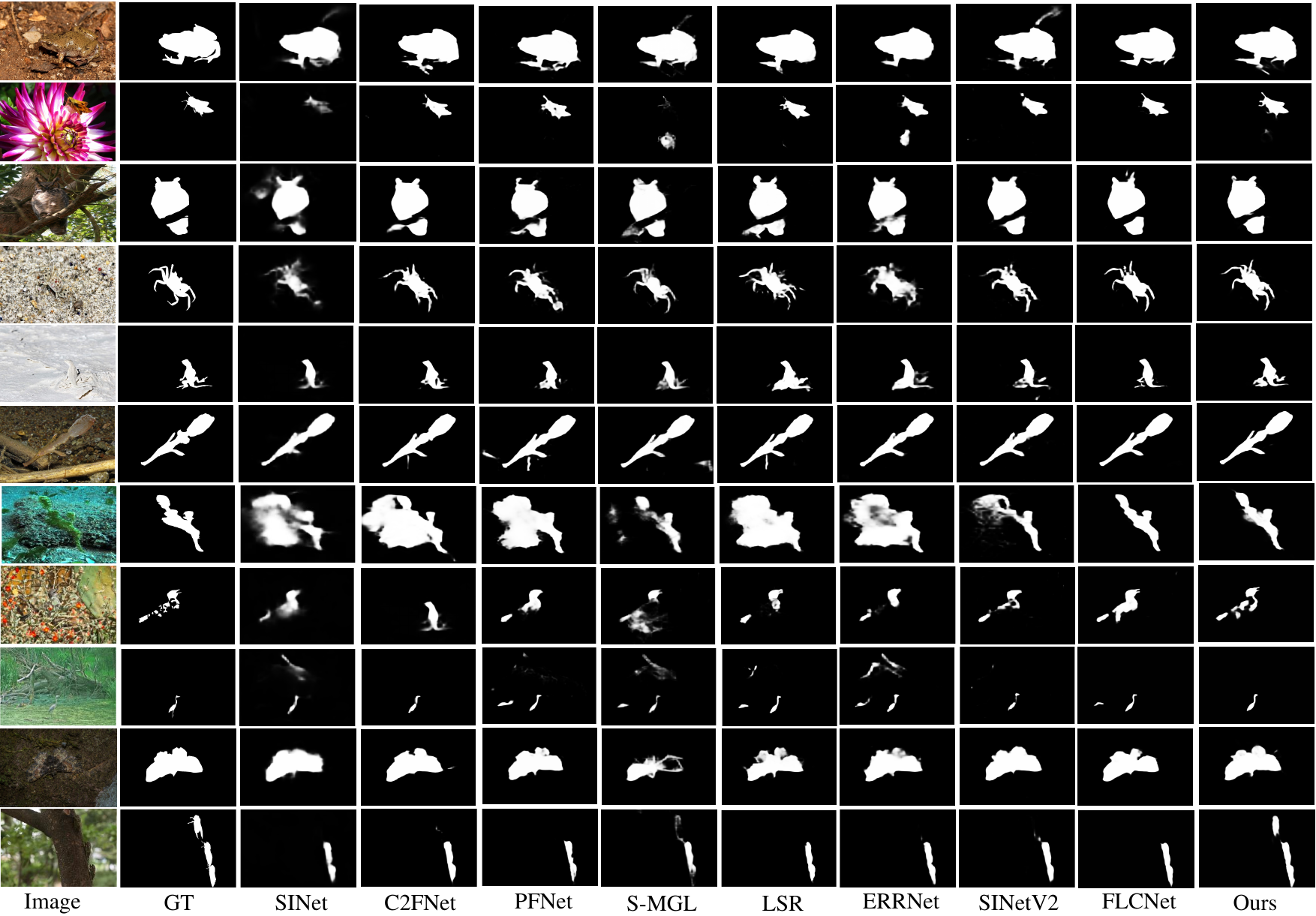}
	\end{overpic}
	\caption{\revII{Qualitative comparison of our MCIF-Net and the baseline methods.}}
	\label{visualization_MCIF-Net1}
	\vspace{5pt}
\end{figure*}

\newcommand{\mywidth}{0.3\textwidth}
\begin{figure*}[!htb]
    \centering
    \includegraphics[width=\mywidth]{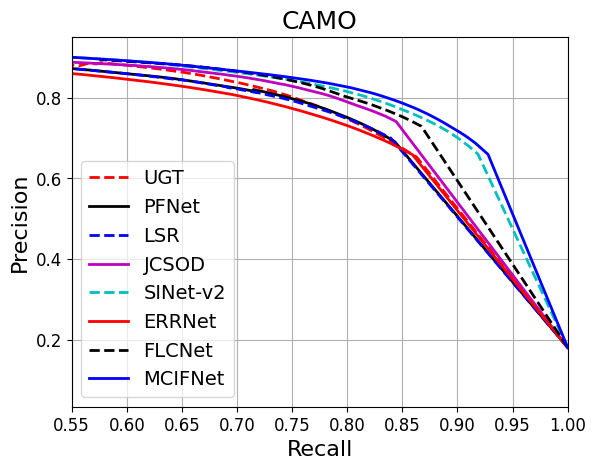}
    \hspace{\fill}
    \includegraphics[width=\mywidth]{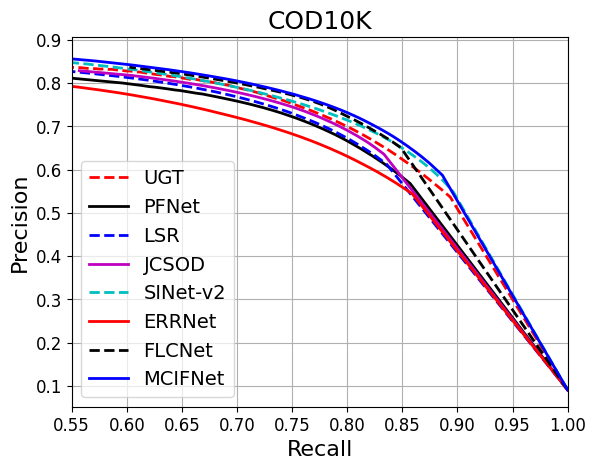}
    \hspace{\fill}
    \includegraphics[width=\mywidth]{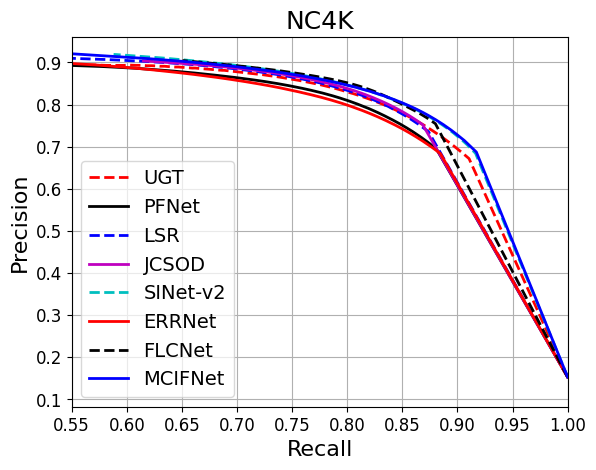}
    \hspace{\fill}
    
    \includegraphics[width=\mywidth]{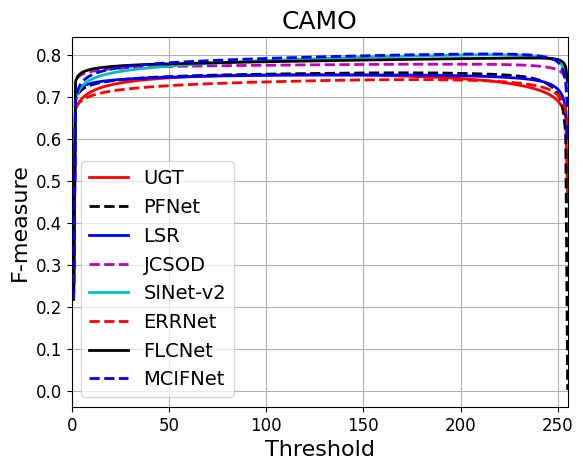}
    \hspace{\fill}
    \includegraphics[width=\mywidth]{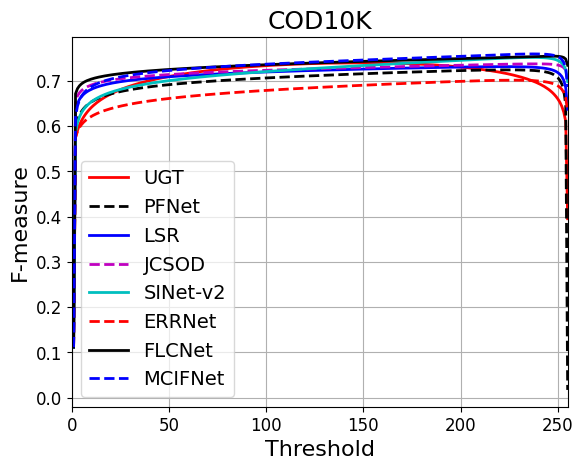}
    \hspace{\fill}
    \includegraphics[width=\mywidth]{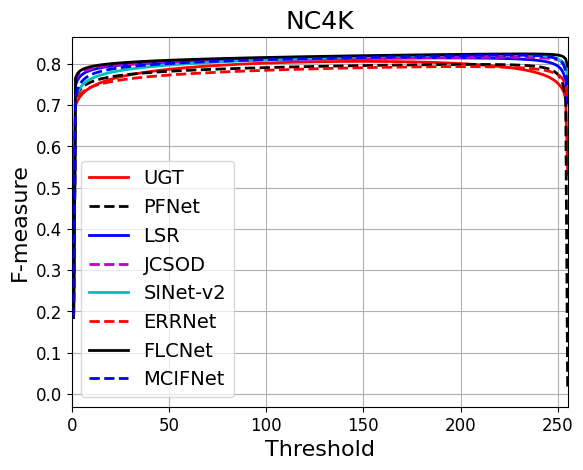}
    \hspace{\fill}

    \vspace{-5pt}
    \caption{The PR curves and F-measure curves of our MCIF-Net and seven state-of-the-art models on challenging COD datasets.}
    \label{pr_curve}
\end{figure*}

   


\textbf{Qualitative evaluation.}
Fig.~\ref{visualization_MCIF-Net1} show the visual comparison of the results given by our MCIF-Net against SOTA models.
As can be observed, MCIF-Net outperforms all competing models and provides the prediction that is the closest to the ground-truth.
In contrast, the competing models are unable to accurately detect camouflaged objects and provide unsatisfactory visual results.
The results, shown in Fig.~\ref{visualization_MCIF-Net1} also demonstrate that our MCIF-Net consistently shows superior performance in different kinds of complex and diverse nature scenes, such as dim light environment, low contrast environment, slender camouflaged object, clutter environment, occlusion condition, \etc~\revxr{Furthermore, an in-depth examination of the Precision-Recall (PR) curves and F-measure curves was conducted to evaluate the performance of our MCIF-Net, rigorously.
As depicted in Fig.~\ref{pr_curve}, the PR/F-measure curve associated with our method consistently maintains a superior position relative to the comparative methods throughout the spectrum, indicating that MCIF-Net achieves enhanced effectiveness both precision and recall, and thereby substantiates its comparative advantage over the existing approaches.} 

\begin{table}[t]
	\centering
	\renewcommand{\arraystretch}{1.2}
	\caption{\small Quantitative results for the ablation studies of MCIF-Net on COD10K and CAMO. For clarity, we use ``B'', ``+DMC'', ``+MIF'', ``w/SE'', and ``w/RFB'' to denote ``Backbone'', ``Backbone+DMC'', ``Backbone+MIF'', ``Backbone+DMC+SE'', and ``Backbone+RFB+MIF'', respectively.
		The best results are in \textbf{boldface}.
		$\uparrow$ indicates the higher the score the better.} \label{compare_Ablation}%

	\renewcommand{\arraystretch}{1.2}
	\setlength\tabcolsep{3pt}
        
	\resizebox{0.5\textwidth}{!}{
        \begin{tabular}{l|cccc||cccc}
			\hline
            & \multicolumn{4}{c||}{\tabincell{c}{COD10K~\cite{fan2020Camouflage}}} &\multicolumn{4}{c}{\tabincell{c}{CAMO~\cite{le2019anabranch}}}
            \\
            \cline{2-9}
            
			 setting &$S_\alpha\uparrow$      &$E_\phi\uparrow$     &$F_\beta\uparrow$      &$M\downarrow$			&$S_\alpha\uparrow$      &$E_\phi\uparrow$     &$F_\beta\uparrow$      &$M\downarrow$
					\\
			\hline
   B  & 0.714  & 0.612  & 0.388  & 0.092 & 0.764  & 0.862  & 0.533  & 0.124  \\
   +DMC   & 0.811  & 0.900  & 0.637  & 0.041 & 0.800  & 0.872  & 0.642  & 0.094  \\
   +MIF   & 0.804  & 0.897  & 0.623  & 0.043 & 0.808  & 0.880  & 0.699  & 0.081  \\
   w/SE   & 0.807  & 0.897  & 0.650  & 0.041 & 0.810  & 0.880  & 0.705  & 0.082  \\
   w/RFB   & 0.817  & 0.902  & 0.680  & 0.037 & 0.809  & 0.876  & 0.722  & 0.077  \\
   \rowcolor{gray!30}
        \ourmodel{}   & \textbf{0.822} &\textbf{0.908}  &\textbf{0.698}  & \textbf{0.034} &\textbf{ 0.823 } & \textbf{0.893}  & \textbf{0.748}  & \textbf{0.071} \\
	\hline
	\toprule
	\end{tabular}}
 \end{table}
			
			

\begin{figure*}[!htb]
	\centering
	\begin{overpic}[width=1\linewidth]{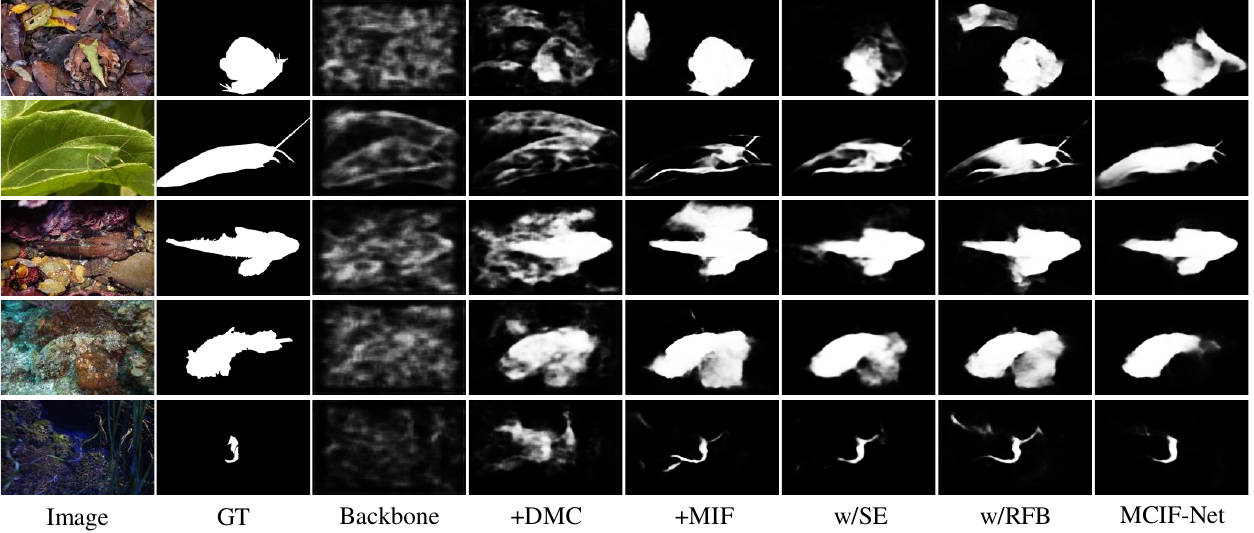}
	\end{overpic}
	\vspace{-15pt}
	\caption{Visual results for the ablation studies of MCIF-Net. For clarity, we use ``+DMC'', ``+MIF'', ``w/SE'', and ``w/RFB'' to denote ``Backbone+DMC'', ``Backbone+MIF'', ``Backbone+DMC+SE'', and ``Backbone+RFB+MIF'', respectively.}
	\vspace{5pt}
	\label{visualization_ablation}
\end{figure*}

\begin{figure}[!htb]
	\centering
	\begin{overpic}[width=1\linewidth]{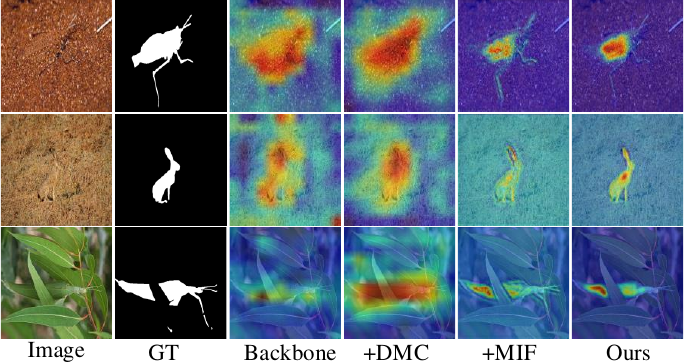}
	\end{overpic}
	\vspace{-8pt}
	\caption{\revII{Visual comparison of attention maps for ablation studies.}}
	\vspace{5pt}
	\label{visualization_ablation1}
\end{figure}

\subsection{Ablation Studies}\label{Ablation_study}
In this section, we perform extensive ablation studies on two benchmark datasets, \ie, COD10K \cite{fan2020Camouflage} and CAMO \cite{le2019anabranch}, to evaluate two proposed modules.
Specifically, two sets of ablation studies are considered. First, we utilize the backbone as the baseline setting, add each proposed module to the backbone as a new setting, and compare two settings to evaluate the effectiveness of the proposed module.
Second, we further demonstrate the effectiveness of DMC and MIF by comparing them with SOTA modules, which are designed for the similar purpose.
Relevant quantitative results are summarized in Table~\ref{compare_Ablation}.
Further visual results are shown in Fig.~\ref{visualization_ablation} and Fig.~\ref{visualization_ablation1}.

\textbf{Effectiveness of DMC}.
We first investigate the effectiveness of DMC module.
The quantitative results, shown in Table~\ref{compare_Ablation}, indicate that ``Backbone+DMC (+DMC)'' outperforms ``Backbone (B)'' in terms of all datasets and evaluation metrics, which demonstrates the effectiveness of our DMC module. 
The DMC is a kind of receptive field modules. To further demonstrate its effectiveness, we compare DMC with the SOTA receptive filed module, RFB \cite{liu2018receptive}. The baseline model, ``Backbone+RFB+MIF (w/RFB)'' denotes a variant of our network, where DMC modules are replaced with RFB modules.
The results, shown in Table~\ref{compare_Ablation}, indicate ``Backbone+DMC+MIF (MCIF-Net)'' outperforms ``Backbone+RFB+MIF (w/RFB)'' in terms of all evaluation metrics, sufficiently demonstrating the advantage of DMC over the cutting-edge receptive field module.
In addition, the visual results, shown in Fig.~\ref{visualization_ablation}, indicate that ``+DMC'' significantly improves the quality of prediction maps in comparison with ``B'' and the full version of our model, ``MCIF-Net'' outperforms ``w/RFB'', demonstrating that DMC is an effective module capable of improving the accuracy of COD.
\revII{Meanwhile, we show the attention maps of different stages in Fig.~\ref{visualization_ablation1}. It can be observed that the backbone can only roughly locate the position of the object. After adding the DMC module, which enlarges the receptive field, the accuracy is improved.}

\textbf{Effectiveness of MIF}.
We then investigate the effectiveness of MIF module. The quantitative results, shown in Table~\ref{compare_Ablation}, indicate that ``Backbone+MIF (+MIF)'' outperforms ``B'' in all evaluations based on different datasets and metrics, demonstrating that MIF is an effective module that improves the performance significantly. To put this into perspective, in the evaluation using COD10K, MIF significantly improves $S_{\alpha}$, $E_\phi$, and $F^{w}_{\beta}$, as well as reduces the $M$ remarkably.
The visual results, shown in Fig.~\ref{visualization_ablation}, indicate that ``+MIF'' outperforms ``B'' in terms of the quality of prediction maps, which confirms our observations in quantitative results and demonstrates the effectiveness of MIF.
In addition, to demonstrate the effectiveness of the proposed interactive attention mechanism, we replace the MIF module in our network with a squeeze-and-excitation (SE) \cite{hu2018squeeze} attention based module, where the low and high-level features are passed through two SE components and then merged together similar to the MIF module. This baseline model is denoted as ``Backbone+DMC+SE (+SE)''.
The results, shown in Table~\ref{compare_Ablation} and Fig.~\ref{visualization_ablation}, indicate that MCIF-Net outperforms ``w/SE'' both quantitatively and qualitatively, demonstrating that our interactive attention design is an effective solution for improving the feature representation and fusion. 
\revII{
Furthermore, the attention maps, shown in Fig.~\ref{visualization_ablation1}, demonstrate that, after adding MIF, the model will focus more on the camouflaged object, which leads to not only more accurate location of the object position, but also more accurate spatial details of the object, such as the edges.}

\begin{table}[t!]
	\caption{\revxr{\small Quantitative results of \ourmodel{} with different backbones on COD benchmark datasets.
 $\uparrow$ indicates the higher the score the better, and vice versa for $\downarrow$.
 The best results are in \textbf{boldface}.
 We evaluate the results using four widely used metrics, including S-measure ($S_\alpha$), E-measure ($E_\phi$), F-measure ($F_\beta$), and MAE ($M$).}}	
	\centering
	\renewcommand{\arraystretch}{1.35}
	\setlength\tabcolsep{5pt}
	\resizebox{0.5\textwidth}{!}{
		\begin{tabular}{l|cccc||cccc}
			\hline
            & \multicolumn{4}{c||}{\tabincell{c}{NC4K~\cite{yunqiu_cod21}}} &\multicolumn{4}{c}{\tabincell{c}{COD10K~\cite{fan2020Camouflage}}} 
            \\
            \cline{2-9}
            
			  Backbone & $S_\alpha\uparrow$      &$E_\phi\uparrow$     &$F_\beta\uparrow$      &$M\downarrow$			
			&$S_\alpha\uparrow$      &$E_\phi\uparrow$     &$F_\beta\uparrow$      &$M\downarrow$ 			 \\
			\hline
 			
     Res2Net & 0.847& 0.912  & 0.770  & 0.047   & 0.822 & 0.908  &0.698  & 0.034   \\
     \rowcolor{gray!30}
     PVT-v2 & \textbf{0.873}  & \textbf{0.932}  & \textbf{0.812}  & \textbf{0.037}  & \textbf{0.845}  & \textbf{0.928}  & \textbf{0.740}  & \textbf{0.028}  \\
    
	\hline
	\toprule
	\end{tabular}}
	\label{compare_data}
\end{table}

\revxr{\subsection{Extension with Transformer}
Transformers, which are initially designed for natural language processing \cite{VaswaniSPUJGKP17}, \revxrI{have been widely applied in computer vision in recent years and achieved significant progress in numerous visual applications.}
Owing to the quintessential attribute of the self-attention mechanism embedded within transformers, the capacity to globally assimilate information has been notably enhanced, thereby facilitating improvements in model efficacy.
In pursuance of augmenting the proficiency of MCIF-Net, a strategic shift was implemented in its architectural foundation—the feature extraction module was transitioned from a Res2Net backbone to the PVT-v2 structure.
Subsequent training regimens administered on the COD benchmark datasets with both Res2Net and PVT-v2 as the underpinning networks have culminated in empirical evidence, as delineated in Table~\ref{compare_data}, validating the superior performance metrics when employing PVT-v2 relative to its Res2Net counterpart across the test datasets.}

\subsection{Failure Cases}
\begin{figure}[!hptb]
\centering
\includegraphics[width=0.99\linewidth]{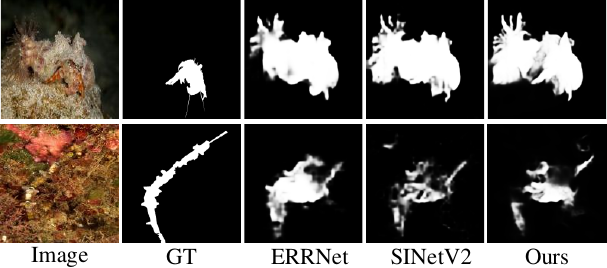}
\vspace{-6pt}
\caption{{Failure cases.} We show the failure cases for three challenging scenarios here.}
\label{fig_failure_cases}
\end{figure}
Fig.~\ref{fig_failure_cases} shows some failure cases of our MCIF-Net.
As can be observed, MCIF-Net fails in some challenging cases, where a large number of camouflaged objects exist and/or the image background is particularly complex.
It is worth noting that the existing cutting-edge model also fails in these challenging cases.
In the future, we will dedicate more efforts to improving the COD accuracy in these cases.

\begin{table}[t!]
	\centering
	\renewcommand{\arraystretch}{1.35}
	\caption{\revII{Quantitative results of polyp segmentation on ETIS. The best results are in \textbf{boldface}. All methods are trained using the same dataset, as in PraNet. $\uparrow$ indicates the higher the score the better, and vice versa for $\downarrow$. We evaluate the results using six widely used metrics, including $\text{Dice}$, $\text{IoU}$, ${F}_{\beta}$, ${S}_{\alpha}$, ${E}_{\phi}$, and $\text{MAE}$.}}
	\vspace{5pt}
	\setlength\tabcolsep{5pt} 
	\resizebox{0.45\textwidth}{!}{
		\begin{tabular}{l|cccccc}
			\hline
			\multirow{2}{*}{{Models}} & \multicolumn{6}{c}{ETIS} \\
			\cline{2-7}    
				&	$\text{Dice}\uparrow$	&	$\text{IoU}\uparrow$	&	${F}_{\beta}\uparrow$	&	${S}_{\alpha}\uparrow$	&	${E}_{\phi}\uparrow$	&	$\text{MAE}\downarrow$	\\
			\hline
				U-Net	&	0.398	&	0.335	&	0.366	&	0.684	&	0.643	&	0.036	\\
				UNet++	&	0.401	&	0.344	&	0.39	&	0.683	&	0.629	&	0.035	\\
				SFA	&	0.297	&	0.217	&	0.231	&	0.557	&	0.531	&	0.109	\\
				ACSNet	&	0.578	&	0.509	&	0.53	&	0.754	&	0.737	&	0.059	\\
				PraNet	&	0.628	&	0.567	&	0.6	&	0.794	&	0.808	&	0.031	\\
			EU-Net	&	0.687	&	0.609	&	0.636	&	0.793	&	0.807	&	0.067	\\
				DCRNet	&	0.556	&	0.496	&	0.506	&	0.736	&	0.742	&	0.096	\\
			\hline
			\rowcolor{gray!30}
			 \textbf{(Ours)} & \textbf{0.717} & \textbf{0.630} & \textbf{0.750} & \textbf{0.808} & \textbf{0.906} & \textbf{0.026} \\
			\hline
		\end{tabular}
	}
	\label{tab:etis}
\end{table}%

\subsection{Application to Polyp Segmentation}\label{sec:Potential_applications}
Our MCIF-Net has shown remarkable performance in COD, which sufficiently demonstrates its practical value. Furthermore, we notice that polyp segmentation can be viewed as a downstream application of COD since they share the same principle with COD, where the polyp has a similar appearance with its surroundings. Polyps are difficult to detect because their color is highly similar to normal tissue, resulting in further deterioration. Referring to the classical polyp segmentation model PraNet~\cite{fan2020pra}, we use the training dataset and test our model on the most difficult polyp dataset ETIS~\cite{silva2014toward}. The parameters used are consistent with those in COD task without optimization. 
We employ six widely-used evaluation metrics, including Dice~\cite{milletari2016v}, IoU, mean absolute error (MAE), weighted F-measure (${ F}_{\beta }^{w}$){~\cite{margolin2014evaluate}}, S-measure (${S}_{\alpha}$)~\cite{Fan2021S-measure}, and mean E-measure (${E}_{\phi}$)~\cite{fan2018enhanced}  to evaluate the model performances.
The results are shown in Table.~\ref{tab:etis}.
As can be observed, our \ourmodel~outperforms the cutting-edge polyp segmentation models in terms of different evaluation metrics, sufficiently demonstrating its superiority in the downstream application of COD.

\section{Conclusion and Future Work}\label{sec:Conclusion}
In this paper, we propose MCIF-Net, a novel deep learning model for the accurate detection of camouflaged objects. MCIF-Net employs the specially-design DMC modules to extract rich context features from a large receptive field and effectively fuses the features using the proposed MIF modules.
Our model provides both the large receptive field and effective fusion, effectively satisfying the demands of accurate COD.
Extensive experiments on \datasets~benchmark datasets demonstrate that MCIF-Net is capable of accurately detecting camouflaged objects and outperforms the SOTA methods.
In addition, our ablation studies verify the effectiveness of two proposed modules, \ie, DMC and MIF, sufficiently.

In the future, we will explore compressing our model for a lightweight one suitable for mobile devices and further improving the efficiency of our model for real-time applications. \revII{In addition, as discussed in Section~\ref{sec:Potential_applications}, our MCIF-Net has shown promising performance in the typical downstream application of COD, \ie, polyp segmentation. 
In the future, we will explore to extend MCIF-Net to more downstream applications of COD, \eg, defect detection.}
\revII{Finally, camouflaged instance segmentation \cite{le2021camouflaged} is a very interesting and significant task. It is also highly relevant to the current COD task, so we will try to migrate the proposed architecture to the camouflaged instance segmentation task in the future.}

\section{Acknowledgments}
This work was supported in part by the National Natural Science Foundation of China under Grant 62201465 and the Fundamental Research Funds for the Central Universities under Grant D5000220213.

{
	\bibliographystyle{IEEEtran}
	\bibliography{reference}
}

\end{document}